# Automated Robotic Moisture Monitoring in Agricultural Fields.

P.Senthil,
Coimbatore Institute of Technology,
Coimbatore, India.
e-mail id-senthillihtnes1994@gmail.com

I.S.Akila,
Coimbatore Institute of Technology,
Coimbatore, India.
e-mail id-Akilasubramaniam.ece.cit@gmail.com

*Abstract:* **Monitoring moisture level of land in a large-scale plantation is tedious. The main objective of this project is to use a robotic kit in collaboration with the on-field moisture sensor circuits, thereby creating an efficient and economical moisture monitoring system. A large agriculture field is divided into smaller grids. Each grid is placed with a moisture sensor. Whenever a sensor reports the soil to be dry, the robot goes to the concerned field for inspection. The path to the concerned field is found by applying Dijkstra's shortest path algorithm on the aerial image of the field. Then the total moisture content of the field is calculated by the robot using suitable image processing algorithms and reported accordingly. For developing and testing this work, a small study field was set up above which a camera was mounted at an appropriate height to capture its aerial view. Thus a prototype for an automated system of monitoring agricultural fields' moisture has been developed through this work.**

## I. Introduction

The objective of our project is to monitor agriculture in an effective way using sensors in integration with a Firebird V robotic kit. The use of sensors in agriculture has been in research for quite a long time. But such a system cannot be a standalone solution for moisture monitoring because the density of sensors required for a real-time scenario would make the implementation meaningless and costly. Our work primarily focuses on integrating the data from these sensors with a robotic kit. The calibration of soil moisture through image processing techniques has been investigated for a while and an image processing algorithm for the same has been developed by us based on previous studies to overcome practical difficulties. The robotic kit replaces human intervention and decreases manual labor. This approach also reduces the density of sensors and is a step towards agricultural automation.

## II. RELATED WORKS

Soil moisture measurement is a challenging problem, where different approaches have been tried out. One such approach was based on L-Band radiometer dual-polarization measurement [1]. The authors of this paper offered a method for the measurement of soil moisture content of a bare soil. This system establishes a method to determine soil dielectric properties based on Radiometer measurements. This system was built upon the self-validated claim that soil roughness can be canceled out using the relationship between V and H polarization. This approach resulted in an indirect method of soil moisture measurement: A simple semi-empirical model (Hp model). Here both active and passive sensor configurations were used. The soil moisture measurement error was very low for bare soil but found to increase considerably for vegetated soil. Though the overall RMSE (Root Mean Square Error) was found to be only 0.043, the study did not propose a generalized method across all soil types and the measurement made in vegetated soil were generally found to be less accurate. Acoustic methods have also been tried out for soil moisture estimation [2]. This method used the known speed of sound in the soil for moisture estimation but fell short of a general system since its approach is limited to the alluvial soil.

Moisture content estimation has also been studied for governmental action for critical applications. The moisture content of a soil is a major factor for risk management in the agriculture sector [3]. This system was designed to assist the Canadian government in framing strategies for agriculture and Agri-food fields. Early assessment of soil moisture content is a very critical factor that can help us predict disaster in agriculture. Here the calibrated Integral Equation Model (IEM) was used to estimate the soil moisture content. The mean average error of this error approach was found to be 7.95%. RADARSAT-2 satellite was used to acquire images of the field under study. When soil moisture estimates were evaluated at a regional scale, mean errors fell to 3.14%. But higher errors were observed for data sets where angles between the RADARSAT-2 look direction and field tillage structures were largest.

The concept of using wireless networks for moisture monitoring has shown promising prospects. A few studies have focused on the basis for forming a wireless sensor network using Zigbee [4, 5, 6]. The Zigbee based wireless sensor networks can measure a variety of physical parameters like weather, soil moisture content, soil temperature, and soil fertility. This paper also discussed the possible issues in the physical layer of the Zigbee. This system was a step towards intelligent farming. A few other studies have gone a step further and discussed a heterogeneous wireless sensor network for measuring different physical parameters, by making use of the emerging technology of Internet of Things (IoT). The advent of IoT into agriculture opened a gateway to bring in the use of data analytics: when the data acquired from the wireless sensor network is uploaded to cloud and processed by a remote PC, many inferences about the field can be made which can be quite useful for agricultural management. In very critical scenarios, a GSM SMS can be sent to the field owner to give a warning about the field status. But one fact that is often overlooked is the density of sensors that are to be used for these applications and the effective cost of implementing such a system. The high density of sensor networks demanded by these ideas makes it difficult to employ them in real time fields.

Analysis of soil properties through image processing is an active area of research [7,8,9,10]. There has been an established correspondence between the color of the soil and its moisture content [11,12] Hence, the use of image processing for measuring soil moisture content has slowly been gaining momentum. Image processing based approaches have been utilized to calculate soil moisture content based on the color characteristics analyzed using RGB and HSV model [13]. In this work, an empirical relationship between extracted features of soil images and soil moisture content was tried out. The soil moisture content was determined using time domain reflectometry and the values were compared with the soil moisture values obtained using image processing and found to have promising results. Our work builds on this work by integrating image processing algorithms with a robotic kit and a low-density wireless sensor network for automated moisture monitoring. Image processing and hardware electronics systems have been experimented in the area of plant disease detection [14, 15, 16]. The study of these systems was helpful in understanding the constraints imposed by a practical agricultural field. Moreover, an image processing based system should be sufficient for assessing the moisture content of a field since we, humans, are able to make judgments about a field's moisture content solely based on visual cues.

The existing systems have very few algorithms for moisture estimation that can be applied satisfactorily for all kinds of soil. Some algorithms work well for bare soil but not for vegetated soil. The number of sensors required increases exponentially when the area of the agricultural land increases. No strong algorithm or system is currently in existence to bring robotics into agriculture. The scope of our work is to bring robotics into agriculture by utilizing image processing algorithms in connection with an economic low-density wireless sensor network.

## III. PROPOSED WORK

The proposed system overcomes the drawback of the existing systems by using moisture sensors on a field with a Fire Bird V ATmega2560 robotic kit.

Moisture sensors alone cannot be independently used to determine the accurate moisture content of an agricultural field. The problem arises because of the low operational range of moisture sensors i.e., a typical moisture sensors' output range indicates the moisture status of only a small vicinity of the area surrounding it. To illustrate this idea more clearly consider a rectangular field in which a few moisture sensors are placed at random locations.

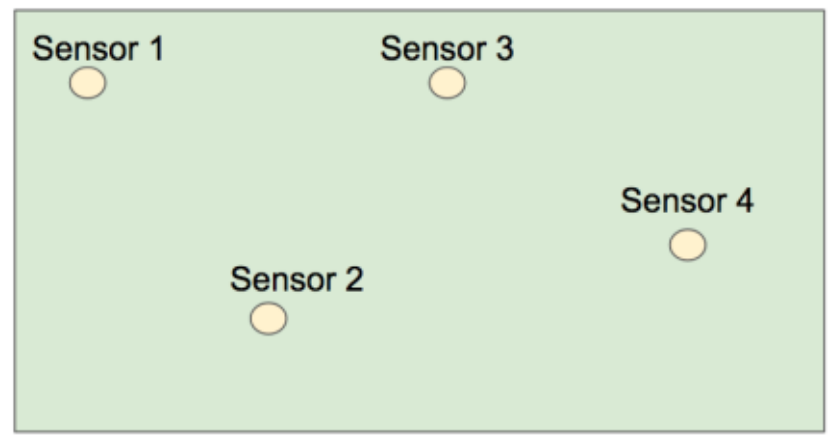

**Fig.1.** The layout of a typical rectangular field with moisture measuring sensors

One straightforward idea is to assume that the number of sensors is large enough to make generalized predictions about the field and take the average of all moisture measurements to be the field moisture measurement. Since the operational range of sensors is very small, the number of sensors is never large enough to make generalized predictions. For example: If sensors 3 in the figure 1 outputs low and all other sensors output high, is a significant portion of the field dry? This question is ambiguous if the distance between the sensors is high.

Another simple approach that could spring up to thought is to model the moisture distribution of the whole field based on the moisture value taken at these sample points. But in order to model the moisture distribution, we have to make assumptions about a generic model for this 3D Moisture distribution function – a function that output the value of moisture at every coordinate (x,y) in the field where x and y are real numbers. This function is not easy to model through a deterministic function since it depends on a lot of environmental and weather factors like direction of sunlight in the area, slope of the land, water flow pattern in the particular field etc.. The only reasonable assumption that can be made about this function is that it is generally smooth and continuous since changes in moisture content in a real-time field are rarely abrupt or discontinuous. But that's not enough to determine this function fully. Though increasing the number of sensors in the field may seem to be a logical option, it's not feasible one for covering a large-scale field in real time usage since the operational range of a sensor is only in the order of a few cms.

After realizing this fact, our focus in the work shifted towards using a robotic kit for moisture monitoring. Rather than using a robotic kit for monitoring moisture independently by inspecting the field throughout the day, it is more efficient to issue commands to the robot to inspect fields on a need basis. This approach follows from the general observation that if all the sensors within a field are outputting high value, there is a high likelihood that the field is not dry and even if one sensor is outputting low, there is a danger that a significant portion of the field might be dry, though it's theoretically hard to prove either case. The underlying assumption behind these claims is that the zones between the sensors will most likely have moisture distribution that is a smooth transition of some form from one sensor point to another. Therefore, it's a safe claim to make and hence, the sparse sensor network can be used in collaboration with a robot to achieve an efficient and reliable moisture monitoring. The output from the sensors serves as a warning for the robot to inspect the concerned field and estimate the actual moisture content of the field.

The only assumption in this approach is that at least one of the sensors present within a field will raise a warning when a significant portion of the field is dry, though a number of those warnings may, in fact, turn out to be false alarms, which will be negated by the robotic kit upon inspection.

A large agricultural field is divided into smaller subfields. A wireless sensor for moisture detection is placed on each subfield. The size of each subfield is taken to be large enough so that the density of sensors required is low, thereby bringing down the cost of the system also. When the moisture of soil in a particular subfield goes below a threshold value, the message is communicated to the robotic kit using Zigbee module. The message from a particular sensor only means that the small vicinity surrounding the sensor is dry but the status of the field as a whole cannot be predicted with very high confidence since only a low-density sensor network is used here. So the Firebird robotic kit goes to the concerned area in the field using Dijkstra's algorithm for shortest path and inspects the whole grid. The whole field is inspected and analyzed using image processing algorithms. Then the message is passed on through Zigbee about the status of the field. Thus a low-density network along with a robotic kit brings about a cost-effective solution.

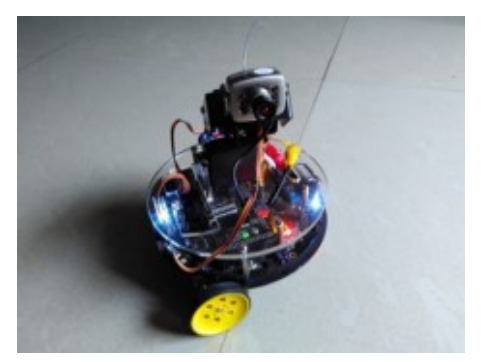

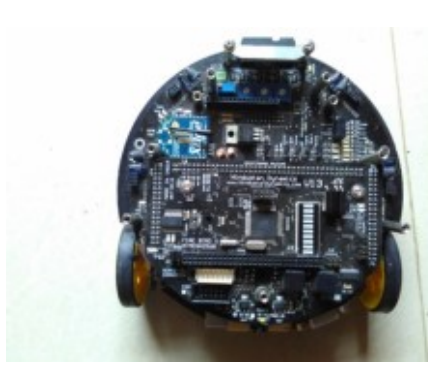

**Fig.2.** Fire bird-V robotic Kit with wireless camera interfaced (left) and top view of fire bird-V tic kit (right).

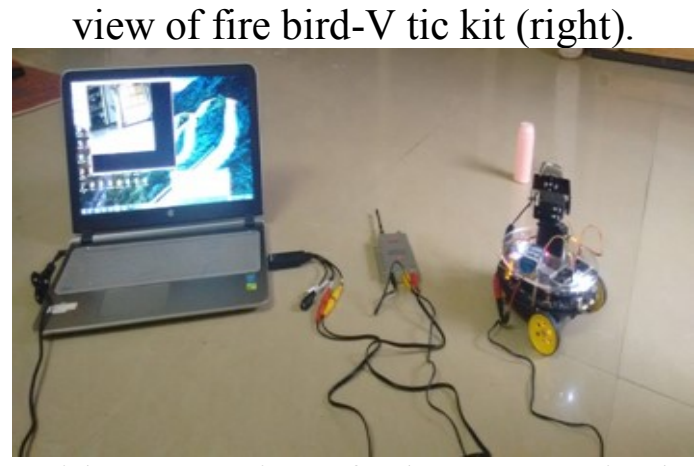

**Fig.3.** Live video streaming of robot camera in the computer

## IV.DIJKSTRA'S ALGORITHM

Dijkstra's algorithm is a popular one for finding the shortest path between two nodes. To develop our work, we initially coded the algorithm for a grid image. Fig 4 shows one of the grid images that was used. Each square is a node and an eight-neighbor connectivity is assumed to construct edges between adjacent nodes. Finally, an unweighted graph is created on which Dijkstra's algorithm is applied. Depending on the robotic movements that are permitted, a suitable neighbor connectivity can be chosen between nodes. Its worth noting that for unweighted graphs like these, Dijkstra's algorithm just degenerates into a breadth-first search algorithm.

The algorithm works by iteratively computing the net distance (number of steps taken to reach that node from the source node) of all unvisited nodes that are adjacent to visited nodes. This process is continued till the end point is reached. Figure 4 shows an image that is divided into grids and the computed net distance assigned to each node as per the algorithm.

Once the destination node is reached, the algorithm finds the shortest path between the source and destination by backtracking the sequence of nodes from the destination node that have consecutively decreasing net distance values till it reaches value starting node (0 represents starting node). Now this list of all nodes traversed will give the shortest path between the source node and the destination node. In this manner, the shortest path for a given source and destination can be obtained for a grid image and one such shortest path is indicated with a blue line in figure 4.

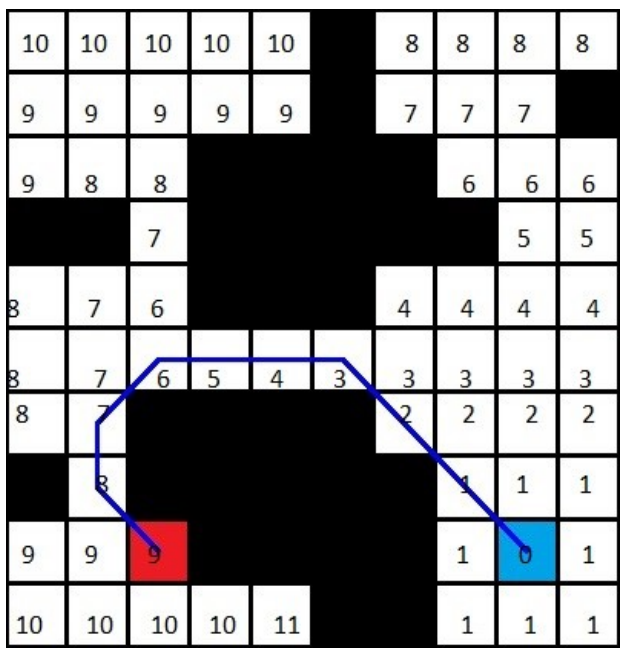

**Fig. 4.** The result of Dijkstra's algorithm

## V.FIELD LAYOUT & PATH PLANNING

For the purpose of this study, a rectangular study field was set up with a dimension of about 200 x 160 cm^2. A rhombus-shaped path (of about 25 cm width) was established in the rectangular field to facilitate the navigation of the robotic kit, thus dividing the larger field into five smaller subfields. Each subfield carried a soil sensor and a Zigbee. A camera was fixed at a height of 8m from the field. The purpose of this camera was to get an aerial image of the field for guiding robotic navigation. The camera at the top is connected to the PC. The image taken by the camera is processed by the PC and it will guide the robotic kit to identify its current position and the path to the destination in the field. In real time scenario, the aerial image will be provided by a satellite/drone terrain mapping.

Once the image of the field is captured, it has to be converted to a form that is suitable for finding the shortest distance between two locations using the Dijkstra's algorithm. Image processing techniques are used for this purpose. Initially, the field alone is masked from the camera image since a few other areas might be visible in its field of view. This is done by finding the biggest contour in the image having black color as its boundary. Once the biggest contour is found it is isolated from the rest of the image and resized to a convenient form. The resized image of the field is shown in figure 5. This step just ensures alignment and crops ROI from the captured image. On the whole, this step makes the image ready for other image processing operations to follow.

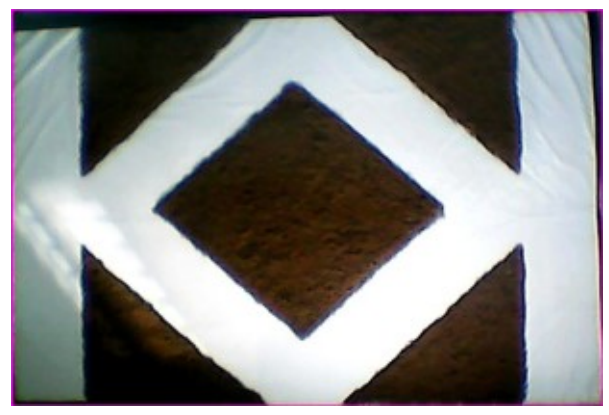

**Fig. 5.** Resized Field Image

As a next step, the image is divided into smaller blocks by drawing horizontal and vertical lines and the image is converted to a grey scale image. Noise, if any present in the image, is removed from the grayscale image using the Gaussian blurring. Once the noise is been removed, it is converted into a binary image as shown in figure 6.

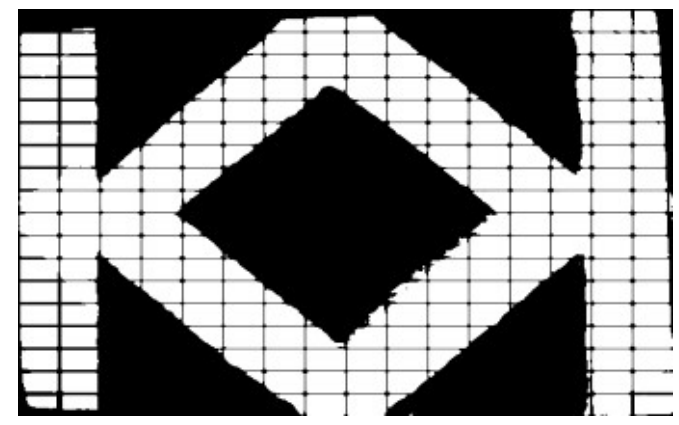

**Fig. 6**. Binary Image of the Grid-Field

After finding the binary image, the square blocks having the white area less than a threshold value are removed from the image. This is because those blocks are too small for a robotic kit to navigate. The image with only navigable blocks is shown in figure 7.

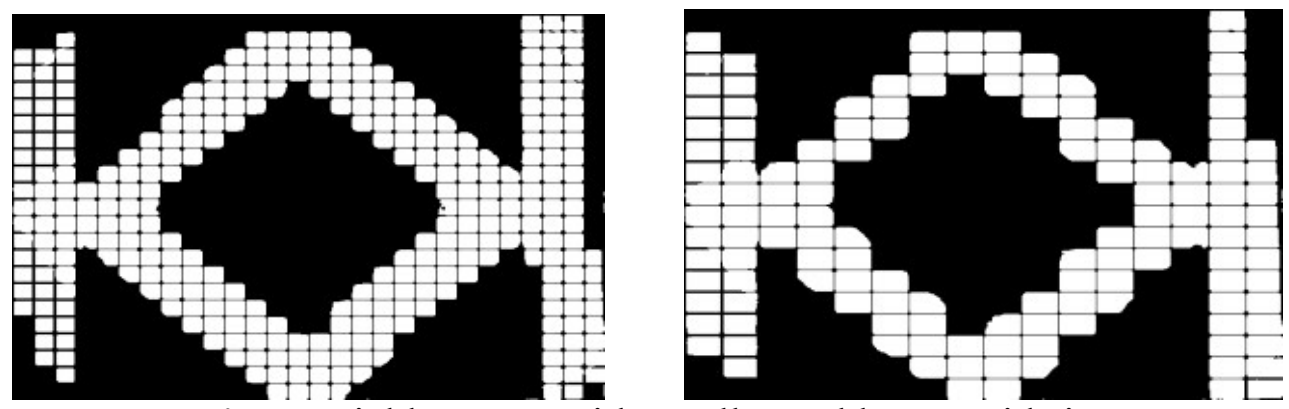

**Fig. 7**. Field Images with smaller and larger grid sizes

As we can see one of the prime factors of this algorithm is the grid size employed in the image. As we decrease the grid size of the image, the image tends to look more like a real-time image rather than a grid-type manipulated image. Decreasing the grid size effectively decreases the smallest distance that can be traversed by the robotic kit, thereby increasing the resolution of its movement. But having a large number of grids also increases the computation time involved. Moreover, very small grid sizes also bring in another problem: if the grid size is very small when compared to the size of the robot, we may find a path between the source and the destination which a robotic kit might not be able to go through because the individual size of one grid might be too small to hold a robotic kit. Too small gird size also facilitates paths with plentiful turns, which should be smoothened for an effective navigation Hence a trade-off is necessary for the grid size.

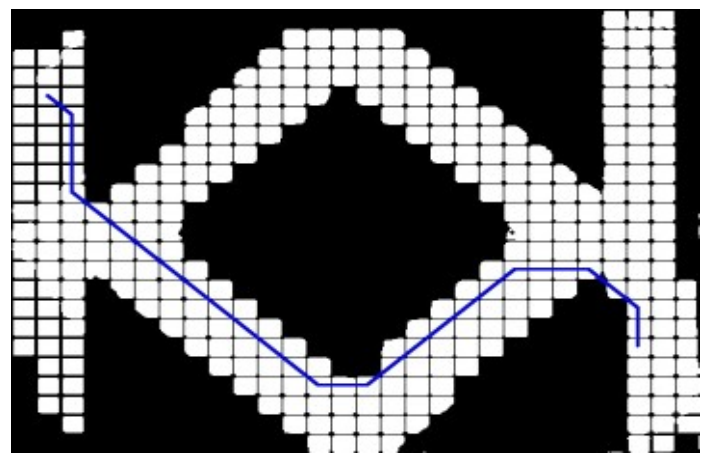

**Fig. 8.** Final Image Showing shortest path

Now the Dijkstra's algorithm is applied to the final image to find the shortest path. With the known position of the robotic kit and the destination node, the path between the two nodes is determined so that the coordinates of the grids through which the kit has to travel can be calculated. This path is drawn using a solid line as shown in figure 8.

When we look at the process involved in transforming the shortest path information in the figure to actual robotic navigation, we need to calculate two important information. At each square block, we have to measure the degree of rotation and the distance to travel after applying the required rotation to move to the neighboring grid. The distance to travel can be easily determined from the calibrated scale of the camera image (ie., the number of cms or mms that each pixel in the image corresponds to in the field setup). For angle calculation, we can utilize the slope of the line joining the two successive square blocks. Thus the commands for robotic navigation can be readily determined and quickly communicated.

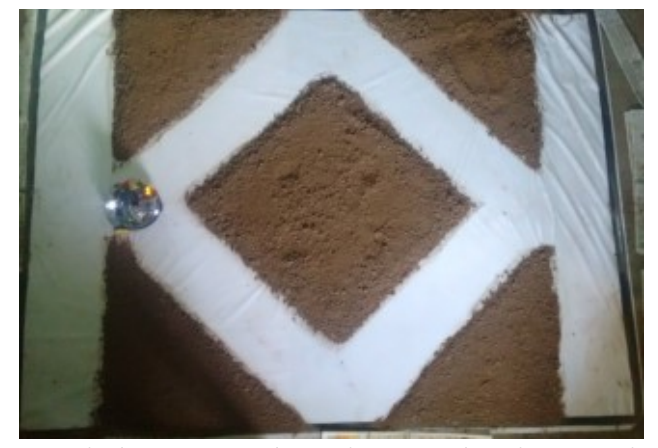

**Fig.9.**Aerial Image showing Robotic Navigation

## VI.CIRCUIT FOR TRANSMISSION AND RECEPTION OF SOIL MOISTURE SENSOR DATA

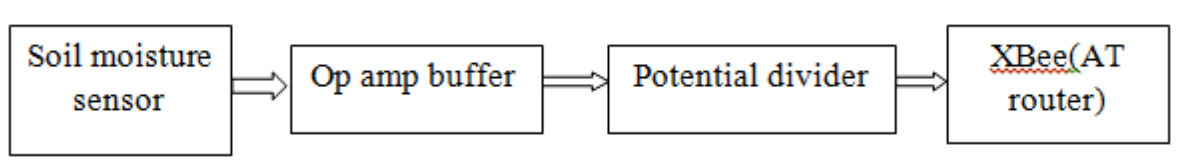

**Fig. 10.** Soil moisture transmission circuit

Each subfield is fixed with a soil moisture sensor, whose output needs to be reported periodically via Zigbee. Figure 10 shows the circuit used for soil moisture sensor data transmission and reception. The internal reference voltage for an XBee's inbuilt ADC (Analog to Digital Converter) is 3.3V but the soil moisture sensor has an output voltage of 5V. Hence there is a need to attenuate the voltage to 3.3V or below. This function is performed by a voltage divider. But the soil moisture sensor used (the hygrometer sensor that's commonly available in the market) has poor output impedance to drive the output load. Hence, an op-amp is used as a buffer before the voltage divider to amplify the current to drive the load. Finally, the data is sent via Xbee which is a router in Application Transfer (AT) mode (Xbee has two modes – AT mode, API mode)

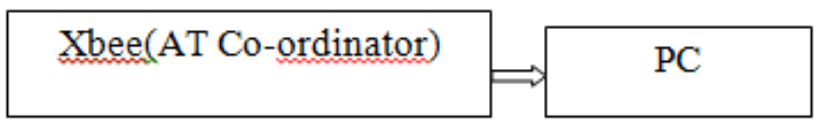

**Fig. 11.** Soil moisture data reception

The data sent is then received by another XBee which is in coordinator AT mode. This XBee is connected to the PC. The received voltage is then multiplied by a suitable factor to get back the original value and then the approximate volumetric soil moisture content at the sensor location is calculated. Based on the calculated moisture content, the PC issues commands to the robotic kit to visit and inspect the field if the moisture content at a particular sensor location is found to be low since there is a high likelihood that the whole of the field might be dry.

## VII. MOISTURE ESTIMATION FROM SOIL IMAGES

If alarm indicating soil dryness is given out by the soil sensor, the robotic kit inspects the concerned subfield and estimates the moisture content of the soil using image processing techniques. Table-I shows the steps involved in the estimation of soil moisture by using the image processing techniques.

TABLE I
Steps for Moisture Estimation

| IMAGE | DESCRIPTION |
|---|---|
| | STEP 1: Field image taken by the robot camera |
| | STEP 2: Masking in HSV domain to remove objects other than sand |
| | STEP 3: Set luminosity constant in the image |
| | STEP 4: Convert the image to a grey scale image. Find the average grey scale value of the image. Get the moisture content of the image through the framed empirical relation |

The measure used here for soil moisture estimation is a volumetric one. Volumetric moisture is simply the ratio of water volume to soil volume. To calculate the moisture content for a given soil image, an empirical relation between average grey scale value of soil image and its volumetric moisture content was determined as follows. Images of soil with known moisture content were taken and their corresponding gray scale values were calculated. A 1st order and 2nd order best fit function for average grey scale value Vs moisture content were calculated. It was found that the 2nd order function gives a better relationship between average grey scale value and percentage of moisture content.

TABLE II
Table showing pre-determined values of volumetric moisture content along with its corresponding grey scale value

| S. No. | Average grayscale value | Percentage of moisture content |
|---|---|---|
| 1 | 98 | 51 |
| 2 | 105.78 | 20.67 |
| 3 | 104.8466 | 23.23 |
| 4 | 104.4055 | 25.25 |
| 5 | 103.6988 | 30.43 |
| 6 | 102.4501 | 31.38 |
| 7 | 102.13877 | 33.78 |
| 8 | 100.2818 | 41.6 |
| 9 | 109 | 10 |

The values in the table are calculated as follows. A completely dry soil (soil sample which was heated and baked and then cooled) of a known quantity was taken and a known quantity of water was added in discrete intervals and imaged continuously. Using the known measurements of soil quantity and water quantity, the volumetric moisture content was calculated and was assigned to the average grey scale value of its corresponding image.

Using the above tabulation, a graph was drawn between the average grayscale value and the volumetric moisture content and an equation relating both was obtained. Figure 12 shows the graph plotted using the above values along with its 1st order and 2nd order best fit function

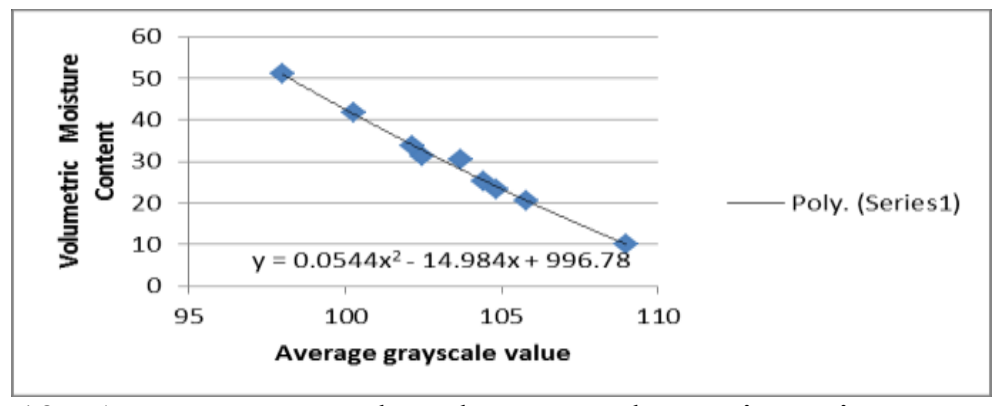

**Fig. 12.** Average grayscale value Vs Volumetric moisture content

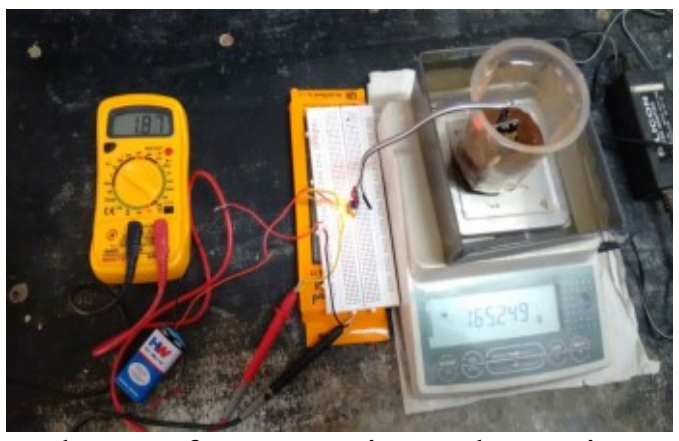

**Fig. 13.** Experimental setup for measuring volumetric moisture content and the corresponding average grayscale image of the soil.

A 2nd order function was found to give lesser MSE (Mean Square Error) than a 1st order function. The 2nd order calibrated equation for the given soil type is

$$y = 0.0544x^2 - 14.984x + 996.78$$

where y is the volumetric moisture content and
x is the average grayscale intensity.

Thus given an image of soil, its average grey scale value could be computed after applying the necessary preprocessing steps and its volumetric moisture content could then be readily calculated by the calibrated equation. Though the method shown here seems a little primitive when compared with today's computer vision algorithms, the empirical relation established here gave very good results for the soil type that was experimented. Thus, once an empirical relation is established for a given soil type, it is found to give decent results for all images of the same type of soil.

It is worth mentioning here that the camera at the top cannot be used to achieve the objective of moisture monitoring since the camera at the top is only a proxy to get a satellite image that can be used for navigating a field. To assess the moisture content of a field directly from the satellite image, there would be a demand for a super high-resolution image that should be free of vegetation. The soil is barely visible in the aerial view of a vegetated field and hence becomes an impossible problem to solve. Moreover, this system has been designed not interact with a satellite (top camera) continuously but only periodically (once in a few days) to update its aerial view of the field to be helpful in robotic navigation.

## VII.CONCLUSION

An automated system of moisture sensing using Firebird-V robotic kit has been developed successfully through this work. The algorithms and methods developed through this work can help in the easy assessment of soil moisture content. This kind of system can be implemented on a large scale field very efficiently. The integration of sensors with a robotic kit proves to be a very efficient system for monitoring moisture. The use of image processing provides a cheap and reliable means of moisture calculations. The algorithms for robotic navigations are very robust and can be used with any kind of aerial images provided the image has the specified resolution. Thus the robotic movement established through this method is smooth and seamless. With more powerful robotic platforms the performance can be further enhanced and this work could run in a field in real-time agricultural fields.

## VIII.FUTURE SCOPE

The whole of this study was carried out in an indoor field under controlled settings. In order to get the system running in real time environment, the following issues need to be addressed.

1. A robust and specialized robotic platform should be developed for the agricultural applications (we used an off-the-shelf robotic platform).
2. Equations for moisture estimation needs to be calibrated for all types for soil.
3. The power drain of a long-range transmitting XBee needs to be minimized.

## IX.REFERENCES

[1] Peng Guo, "A New Algorithm for Soil Moisture Retrieval With L-Band Radiometer", Applied Earth Observations and Remote Sensing, IEEE Journal, Volume-6, Issue-3, pp. 1147 – 1155, February 2013.

[2] Adamo.F, "An acoustic method for soil moisture measurement," Instrumentation and Measurement, IEEE Transactions, Volume-53, Issue-4, pp. 891 – 898, August 2014.

[3] McNairn.H, Res. Branch, "Monitoring Soil Moisture to Support Risk Reduction for the Agriculture Sector Using RADARSAT-2," Selected Topics in Applied Earth Observations and Remote Sensing, IEEE Journal, Volume-5, Issue- 3, pp. 824 - 834, June 2012.

[4] Deng Xiaolei, "Development of a field wireless sensors network based on ZigBee technology," World Automation Congress (WAC), volume-3, pp. 375-379, June-2013.

[5] Kalaivani.T, "A survey on Zigbee based wireless sensor networks in agriculture," Trendz in Information Sciences and Computing (TISC) IEEE Journal, volume-3, pp. 85-89, December 2012.

[6] Xufeng Ding, "Environment monitoring and early warning of facility agriculture based on heterogeneous wireless networks," IEEE Conference on Service operation and logistics and Inference Publications, volume-3, pp. 307-310 July 2013

[7] Viscarra Rossel.R.A, Cattle.S.R, OrtegaA and Fouad.Y (2009), "In situ measurements of soil colour, mineral composition and clay content by vis-NIR spectroscopy", Soil Science Society of America Journal, Vol.15, Issue-4, pp. 253–266.

[8] Rafael C.Gonzalez and Richard E.Woods. (2010), "Digital Image Processing," IEEE Journal of Trends in image processing, Issue-3, pp. 224-257.

[9] Mrutyunjaya.R, Dharwad.M, Toufiq.A and Badebade.K (2014), "Estimation of Moisture Content in Soil Using Image Processing," International Journal of innovative research, Vol.3, Issue- 4, pp. 49-58.

[10] Konen.M.E, Burras.C.L. and Sandor.J.A (2012), "Organic carbon, texture and quantitative colour measurement relationships for cultivated soils," SoilScience Society of America Journal, Vol.6, Issue-5, pp. 823–830.

[11] Post.D.F, Levine.S.J, Bryant.R.B, Mays.M.D, Batchily.K, Escadafal.R and Huete.A.R (2011) "Correlations between field and laboratory measurements of soil colour," Soil Science Society of America Journal, Vol.14, Issue-6, pp. 35–50.

[12] Bullock.P, and Thomasson.A.J (2011), "Measurement and characterization of macroporosity by image analysis and comparison with data from water retention measurements," IEEE Journal of Soil Science, Vol.2, Issue-5, pp. 391–413.

[13] Persson.M, (2010), "Estimating surface soil moisture from soil color using image analysis," Vadose Zone Journal, Vol.4, Issue-3, pp. 119–122.

[14] Frode Urdal, Trygve Utstumo & Jan Kare Vatne, "Design and Control Of Precision Drop-On-Demand Herbicide Application in Agricultural Robotics," 13th International Conference on Control Automation Robotics & Vision (ICARCV), pp. 1689-1694, Oct 2014.

[15] Baquero.D, (2014), "An image retrieval system for tomato disease assessment," IEEE Journal of Image Signal Processing and Artificial Vision (STSIVA), Vol.2, Issue-4, pp. 1-5, 17-19.

[16] K.P.Park Chang, A.Zakaria & N.Yusuf, "Analysis and Feasibility Study Of Plant Disease Using E-Nose," IEEE Conference on Control system computing and engineering Publications pp. 58-63, 2014

[17] A.Kumar, M.O.Arshad & S.Mathavan, "Smart Irrigation Using Low-Cost Moisture Sensors and Xbee-based Communication," IEEE Conference on Global Humanitarian Technology Publications pp. 333-337, Oct 2014.